\documentclass[10pt,twocolumn,letterpaper]{article}

\usepackage{cvpr}
\usepackage{times}
\usepackage{epsfig}
\usepackage{graphicx}
\usepackage{amsmath}
\usepackage{amssymb}
\usepackage{epstopdf}
\usepackage{ulem}

\usepackage[breaklinks=true,bookmarks=false]{hyperref}

\cvprfinalcopy 

\def\cvprPaperID{****} 

\begin{document}

\title{Attribute-guided Feature Extraction and Augmentation Robust Learning for Vehicle Re-identification}

\author{Chaoran Zhuge\textsuperscript{1} , Yujie Peng\textsuperscript{2} , Yadong Li\textsuperscript{2} , Jiangbo Ai\textsuperscript{3}, Junru Chen\textsuperscript{4}\\
University of Science and Technology of China\textsuperscript{1} \\
School of Computer Science and Engineering, Beihang university\textsuperscript{2} \\
University of Electronic Science and Technology of China\textsuperscript{3}\\
Xidian University\textsuperscript{4}
}

\maketitle

\begin{abstract}
Vehicle re-identification is one of the core technologies of intelligent transportation systems and smart cities, but large intra-class diversity and inter-class similarity poses great challenges for existing method. In this paper, we propose a multi-guided learning approach which utilizing the information of attributes and meanwhile introducing two novel random augments to improve the robustness during training. What’s more, we propose an attribute constraint method and group re-ranking strategy to refine matching results. Our method achieves mAP of 66.83\% and rank-1 accuracy 76.05\% in the CVPR 2020 AI City Challenge.
\end{abstract}

\section{Introduction}

In recent years, the development of technology in the field of computer vision and the breakthrough of technology in the field of Internet of Things promote the realization of smart city concept. As important objects in smart city applications, vehicles have attracted extensive attention, a lot of researches about vehicles has been carried out, such as vehicle detection, vehicle tracking, fine-grained vehicle type recognition, etc. 

As a frontier and important research topic, vehicle re-identification (ReID)\cite{hsu2019multi} also caused more and more attention in research area, the purpose of vehicle ReID is to identify the same vehicle through multiple non-overlapping cameras. A vehicle ReID system can quickly get the location and time of the target vehicle in the city. Vehicle ReID technology is crucial to the future development of the Internet of things, as well as the construction of intelligent transportation system and smart city\cite{cityflow}.

There are two main challenges of vehicle ReID, intra-class difference and inter-class similarity. Intra-class difference is mainly caused by viewpoint variation, what’s more, background clutters, resolution and illumination also have great influence. Inter-class similarity is in reflected images of different vehicle may look very similar. Vehicles produced by the same or different manufacturers can have similar colors and shapes.

In this paper, our proposed method is focusing on extracting robust features for vehicle ReID task, Finally, ensemble and re-ranking is also used to refine the results.

In summary, our contributions are:
\begin{itemize}
\itemsep=-5pt
    \item\small{we proposed a method utilizing attributes information for vehicle ReID.}
    \item\small{we introduce random shrink and background substitution augments to improve the robustness of model when the quality of images is largely different.}
    \item\small{we introduce an attribute constraint method and group re-ranking strategy to get more accurate matching results. }
    \item\small{Our vehicle ReID method achieves mAP of 66.83\% and rank-1 accuracy 76.05\% in the CVPR 2020 AI City Challenge.}
\end{itemize}

\begin{figure*}
\begin{center}
\fbox{\includegraphics[width=0.8\textwidth]{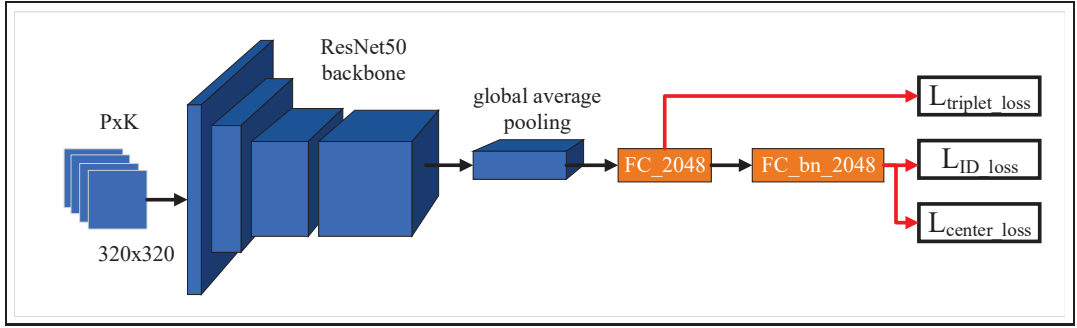}}

\end{center}
   \caption{General architecture of our ResNet50 feature extractor. We removed the last stage down-sampling operation and use IBN to replace BN after the first 1x1 conv on the residual branch. $F_{bn}$ refers to the features after BNNeck. The structure of other feature extractors is basically the same as the ResNet50 feature extractor, with only slight differences. For Densenet161 feature extractor, we keep the last spatial down-sampling operation in the network. For HRnetw18c feature extractor, We use BN instead of IBN after the first 1x1 conv on the residual branch. All feature extractors output $F_{bn}$ feature as an apparent feature. }
\label{fig:short}
\end{figure*}

\section{Related Work}
Similar to face recognition and pedestrian ReID, vehicle ReID also needs to use the trained CNN network to extract the global information of the vehicle images, and make feature similarity judgment in the embed layer; as the CNN network continues to develop, some excellent network structure has been widely used \cite{resnet}\cite{inception}\cite{ibn}\cite{efficientnet}\cite{deephigh}; in the field of face recognition, the softmax Loss based on margin \cite{cosface}\cite{additivemargin}\cite{arcface} increases the distance between intra-classes and compresses the distance between inter-classes, thereby greatly improving the performance of face recognition; in the field of pedestrian ReID, many methods \cite{perceive}\cite{eanet}\cite{circle} are also instructive to explore the vehicle ReID method;

In the vehicle ReID task, Liu et al \cite{ram}used the method similar to PCB in pedestrian ReID to cut the channels/width\-/height dimensions of the feature map, combined with the vehicle's feature information, supervised the network training, and concatenate the features output by each branch network as the final feature; Wang et al \cite{orientation} present 20 fixed key points for the vehicle, and combining the key points to strengthen the extracted special features to improve the representativeness of the feature; Chu et al \cite{chu2019vehicle} combined the orientation information of the vehicle, and processed the vehicles with the same orientation and the vehicles with different orientations separately to reduce the interference of the orientation on the results ; He and others \cite{he2019part} use the detection model to obtain areas such as car lights and annual inspection marks, and then enhance the characteristics of these areas to improve the characterization ability of the features; Lou and others \cite{lou2019veri} use GAN to generate some hard negative vehicles with the same brand but different details in the car, and then use these difficult samples to train the network; Sochor and others \cite{sochor2016boxcars} use the vehicle's 3d box and perspective information to assist the network in extracting features.

Among other techniques, Luo et al \cite{luo2019strong}. Proposed some techniques when designing or training networks, including designing BNNeck, applying triplet loss and classification loss to different feature vectors, and strategies for learning rate warm up. The combination of ID loss, center loss and triplet loss, Random Erasing Augmentation and other methods are very effective. At the same time, the re-rank method proposed by Zhong et al \cite{zhong2017re}. Utilizes the mutual information between query and gallery to further optimize the performance of re-identification.

\section{Proposed Approach}
In this section, we introduce our proposed vehicle ReID method. Firstly, five feature extractors are applied with different model architectures to get robust features. Those feature extractors are trained with cross-entropy loss, triplet loss and center loss. Finally, a new re-ranking method by using query clustering and attribute constraint is adopted to refine the final results.
\subsection{Feature Extraction}\label{featex}
The based feature extractor uses ResNet50 network which is initialized with pre-trained parameters on ImageNet \cite{deng2009imagenet}. We removed the last stage down-sampling operation \cite{luo2019bag} because higher resolution feature maps enrich fine-grained features. IBN-Net \cite{ibn} shows that combining Instance Normalization \cite{ulyanov2016instance}  and Batch Normalization (BN) \cite{ioffe2015batch} with an appropriate manner improves both learning and generalization capacities. For each block of ResNet50, we use IBN to replace BN layer after the first 1x1 conv on the residual branch. The 2048-dim features $f$ are extracted by a fully-connected layer after ResNet50 network. Because of the target of ID loss and triplet loss are inconsistent in the embedding space, we apply a BNNeck \cite{luo2019bag} with zero bias after fully-connected layer to get the normalized feature $f_{bn}$. In the training stage, $f_{bn}$ are used to compute ID loss, and $f$ are used to compute triplet loss and center loss.

To get more distinguished features, we build five feature extractors based on the baseline by using different model architecture. we built a ResNet50 feature extractor as the backbone. Besides, We built a HRnetw18c \cite{sun2019high} feature extractor and a Densenet161 \cite{huang2017densely} feature extractor. Then, an Attribute-Guided Network is trained for getting detailed features. In order to get a robust features against background noise, A Densenet161 feature extractor trained with background substitution is applied. Five feature extractors are trained separately. In the test phase, we concatenate all apparent features together as the final appearance signature, and use the concatenated features to compute the euclidean distances between query images and gallery images.

\textbf{Random shrink:} After cropped with bounding box, the average size of object images varies largely. Therefore, we adopt random shrink augment to improve the performance under this situation. For object with size larger than target resize shape, we generate a random number between 0.4 and 0.6 as the scaling factor. Then, we randomly scale down objects by the scaling factor with a probability of 0.5, and resize it to the input size. This operation can significantly improve the performance on small objects in test set.

\textbf{Background substitution:} In our experiments, training with simulation dataset helps to improve the performance of our model greatly. Inspired by this, we trained a PSPnet \cite{zhao2017pyramid} to get the segmentation mask of vehicle identity and use it to split vehicle and background. Before training, background in images are randomly replaced from other input with a probability of 0.5.

\textbf{Feature ensemble:} To get more general and accurate features of vehicles in testing phase, we ensemble models by concatenating the features generated by ResNet50, Densenet161 with background substitution, HRnetw18c and Attribute-Guided Network(AGN) with corresponding weights as the appearance signature.

\textbf{Loss function:} Cross-entropy loss, triplet loss and center loss are used during training phase. Given an image, we denote $y$ as truth label of ID and $p_i$ as ID prediction logits of class i. We use feature $f_{bn}$ to get ID prediction logits through softmax operation. The cross entropy loss is computed as: 
\begin{equation}
    L_{\text {ID}}=\sum_{i=1}^{N}-q_{i} \log \left(p_{i}\right)\left\{\begin{array}{l}q_{i}=0, y \neq i \\ q_{i}=1, y=i\end{array}\right..
\end{equation}

Center loss are used to learn a center for deep features of each class and penalizes the distances between the deep features and their corresponding class.We use feature $f_{bn}$ to compute center loss. The center loss function is formulated as:
\begin{equation}
    L_{\text {center}}=\frac{1}{2} \sum_{j=1}^{B}\left\|\boldsymbol{f}_{t_{j}}-\boldsymbol{c}_{y_{j}}\right\|_{2}^{2},
\end{equation}
where $y_j$ is the label of the $j$th image in a mini-batch. $C_{y_j}$ denotes the $y_j$ th class center of deep features. B is the number of batch size. 

Simultaneously,we use feature $f$ before BNNeck to compute triplet loss. The triplet loss function is formulated as:
\begin{equation}
    L_{\text {triplet}}=\left[d_{p}-d_{n}+\alpha\right]_{+},
\end{equation}
where $d_p$ and $d_n$ are the distances of positive pair and negative pair in the feature space. The $\alpha$ is the margin of triplet loss, and $[z]_+$ equals to max(z,0). In our experiments,we set $\alpha$ to 0.5. 

Finally,the total loss for our feature extractor is:
\begin{equation}
L=L_{\text {ID}}+L_{\text {triplet}}+\beta L_{\text {center}},
\end{equation}
Where $\beta$ is the weight of $L_{\text {center}}$, In our experiments, $\beta$ is set to 0.0005. 

\begin{figure*}
\begin{center}
\fbox{\includegraphics[width=0.8\textwidth]{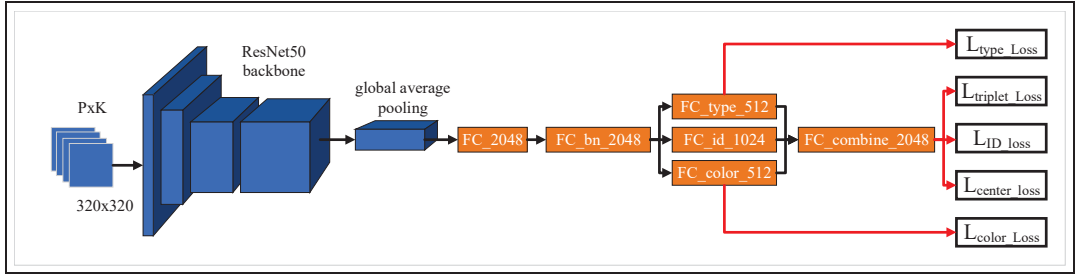}}

\end{center}
   \caption{General architecture of our Attribute-Guided Network. This extractor structure is basically same as the ResNet50 feature extractor, but after BNNeck we add three new fully-connected layer for type, color and id attributes. finally, we concatenate three features after three fully-connected layer and add two cross-entropy losses to compute color loss and type loss. This feature extractor outputs $f_{id}$ feature as an apparent feature.}
\label{fig:attr}
\end{figure*}

\subsection{Attribute-Guided Network}
Attributes like car type, color, etc., are robust against the variations of vehicle appearance, which mainly caused by occlusion and different viewpoints.  Inspired by \cite{ahmed2015improved}, we proposed an Attribute-Guided network. As shown in Fig.~\ref{fig:attr}. A general feature map $F$ are generated by a ResNet50 backbone. Then, $F$ is fed into a master branch to generate a ReID feature  $f_{id}$ and two attribute branches to get attribute-guided feature$f_{type}$ and $f_{color}$ for attribute classification task. The final concatenated feature $f_c$ is regarded as the feature signature for vehicle ReID task.

In type and color classification branches, cross entropy loss is used for multi-task learning. And as same as feature extractors in \ref{featex}, the loss $L_{reid}$ is combined with cross entropy loss, center loss and triplet loss in ReID branch. 

The overall loss function $L_{attr}$ of Attribute-Guided Network can be formulated as:
\begin{equation}
    L_{attr}=L_{reid} + \alpha L_{type} + \beta L_{color},
\end{equation}
where $\alpha$ and $\beta$ are corresponding weights of $f_{type}$ and $f_{color}$. In our experiment, we set both $\alpha$ and $\beta$ to 1.

\subsection{Post-processing}
During the inference phase, two post-processing strategies are used to improve the ReID performance.

\textbf{Attribute constraint:} In order to better separate cars in different type and color in ReID task, We manually label type and color attributes for all Ids in track2 dataset and train a color classification network and a type classification network respectively to label the attributes of benchmark vehicles. And a constraint strategy is utilized on euclidean distance between query and gallery. After the constraint being adopted, the new distance $d$ between query $q_i$ and gallery $g_j$ can be computed as:
\begin{equation}
    d(i,j) =
    \begin{cases}
        d_{old}(i,j) + \delta_t \quad \quad \quad T_{i}\ne T_{j}\,and\, C_i = C_j \\
        d_{old}(i,j) + \delta_c \quad \quad \quad T_{i}= T_{j}\,and\, C_i \ne C_j \\
        d_{old}(i,j) + \delta_c + \delta_t \quad  T_{i}= T_{j}\,and\, C_i = C_j \\
        d_{old}(i,j)\quad \quad \quad \quad \quad \ otherwise \ ,
    \end{cases}
\end{equation}
where $T_i$ and $ T_j$ are car types of $q_i$ and $g_j$, $C_i$ and $C_j$ are colors of $q_i$ and $g_j$, $d_{old}(i,j)$ means the euclidean distance between $q_i$ and $g_j$, $\delta_t$ and $\delta_c$ are two punish-values.

\textbf{Group Re-ranking:} Due to different quality and pose variation of a query, getting the accurate representation is hard from only one image. To solve this problem, we group the queries with Euclidean distance lower than $\theta$, and set the mean feature of the group as their representation. Meanwhile, galleries are clustered with same tracklet. Our re-ranking strategy is based on the query clusters and gallery clusters.

Given $i$th query $q_i$ with group mean feature $c_i$ and $j$th gallery $g_j$ with feature $f_j$. we regard $g_j$ as the similar identity of $q_i$ if the Euclidean distance between $q_i$ and $g_j$ is lower than $\theta$, and then galleries with the same tracklet of $g_j$ are inserted after $g_j$ in the match sequence of $q_i$.

\section{Experiments}
\subsection{Dataset}
The AICITY2020 track2 dataset have two sub-data sets: AIC20 ReID dataset and AIC20 ReID Simulation dataset. AIC20 ReID dataset contains 56,277 images of 666 identities,which all come from multiple cameras placed at multiple intersections.The training set has 36,935 images which come from 333 vehicle identities. The test set consists of 18,290 images which belong to the other 333 identities. The rest of 1,052 images are used as query set. On average, each vehicle has 84.50 image signatures from 4.55 camera views. For the training set,each image has camera ID and vehicle ID ground truth. In addition, we also manually  marked type ID, color ID, bounding boxes and segmentation ground truth for each image in the training set.AIC20 ReID Simulation dataset is generated by VehicleX, which is a publicly aviablable 3D engine.   It has 192,150 images of 1,362 identities. Each image has camera ID,vehicle ID,color ID,type ID and the other ground truth. In our experiment, we add all images of them to the training set.
\subsection{Implement Details}
In training phase of all feature extractors, all input images are resized to 320 $\times$ 320 and padded with 10 pixels on image border, then images are randomly cropped to 320$\times$320 again. Also, we apply random shrink and random erase with a probability of 0.5. Each feature extractor is trained for 120 epochs by Adam optimizer with weight decay of 5e-4. Warm-up learning strategy\cite{luo2019bag} are used in training, learning rate is linearly increased from 3.5e-5 to 3.5e-4 in the first 10 epochs, then divided by 10 at 40 epochs and again at 70 epochs. For ResNet50 and HRnet based feature extractors, we randomly select 64 identities and sample 4 images for each identity. For Densenet161 based feature extractors, we only select 24 identities.

\begin{figure*}
\begin{center}
\fbox{\includegraphics[width=0.8\textwidth]{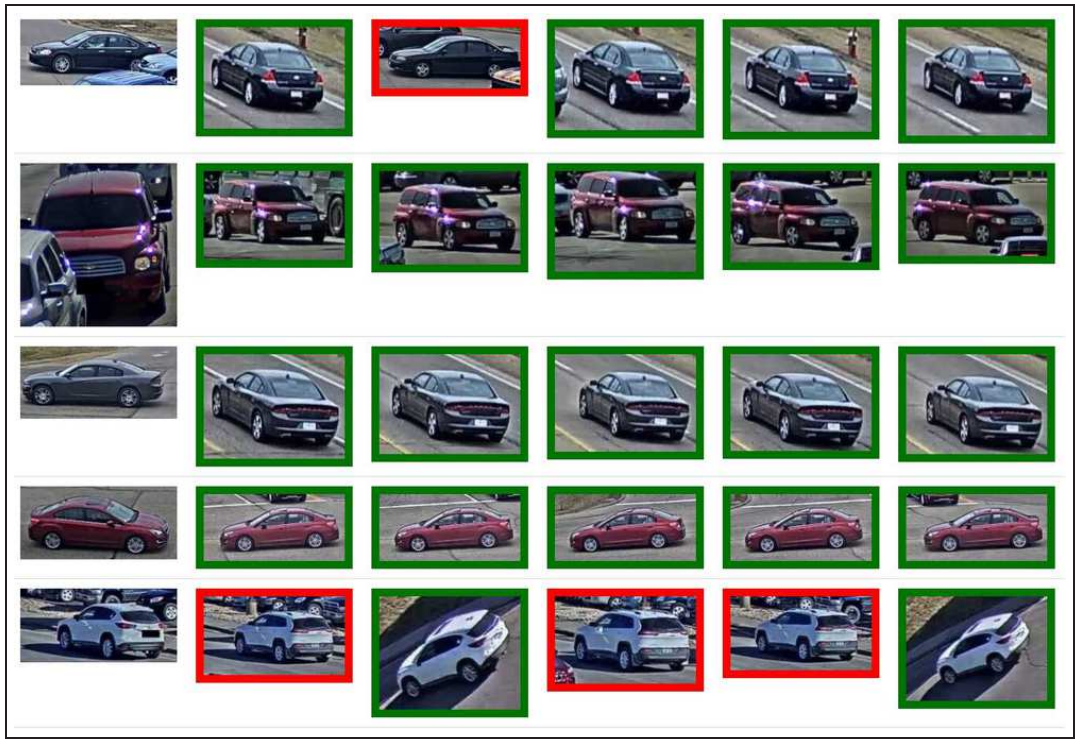}}

\end{center}
   \caption{example ranking results of our method.}
\label{fig:result}
\end{figure*}

\begin{table}
\begin{center}
\begin{tabular}{|c|c|c|}
\hline
{Rank} & {Team ID} & {Score} \\
\hline
1     & 73    & 0.8413 \\
    2     & 42    & 0.781 \\
    3     & 39    & 0.7322 \\
    4     & 36    & 0.6899 \\
    5     & 30    & 0.6684 \\
    6     &{\textbf{44(Ours)}} & 0.6683 \\
    7     & 72    & 0.6668 \\
    8     & 7     & 0.6561 \\
    9     & 46    & 0.6206 \\
    10    & 81    & 0.6191 \\
\hline
\end{tabular}
\end{center}
\caption{Performance Evaluation of Track2.}
\label{finalresult}
\end{table}

\begin{table}
\begin{center}
\begin{tabular}{|c|c|}
\hline
{model} & {mAP}\\
\hline
Res50     & 0.614   \\
Res50+AGN     & 0.638 \\
Res50+AGN+Dense161BS     & 0.657  \\
Res50+AGN+Dense161BS+Dense161     & 0.662  \\
Res50+AGN+Dense161BS+Dense161+HRnet     & \textbf{0.668}  \\
\hline
\end{tabular}
\end{center}
\caption{results of model ensemble. Dense161BS mean Dense-net161 feature extractor trained with background substitution augment.}
\label{ensemble}
\end{table}

\subsection{Performance Evaluation of Challenge Contest}
We report our challenge contest performance of the track2: City-Scale Multi-Camera Vehicle Re-Identification. In Track2, we rank number 6(team ID 44) among all the teams with the mAP of 0.6683, as can been in Table.~\ref{finalresult}. We show some ranking results in Fig.~\ref{fig:result} produced by our method. And comparisons with model ensemble is shown in Table.~\ref{ensemble}.

\section{Conclusion}
In this paper, we introduce a multi-feature extractor vehicle Re-ID system for learning a robust appearance feature for vehicle Re-ID. Based on a mature vehicle Re-ID system, we use pre-processing discussed above to get robust features. Then, we adopt the feature ensemble in our system to get powerful feature representations. After that, we use group re-ranking and attribute constraint to reduce the number of irrelevant gallery images. Finally, our proposed system rank number 6(team ID 44) among all the teams with the mAP of 0.6683 for City-Scale Multi-Camera Vehicle Re-Identification.

{\small

\bibliographystyle{ieee_fullname}
}

\end{document}